\documentclass{article}

\usepackage{amsmath,amsfonts,amssymb,times,graphicx,natbib,algorithm,algorithmic,hyperref}

\usepackage[accepted]{whi2018}

\icmltitlerunning{AI in Education needs interpretable machine learning: Lessons from Open Learner Modelling}

\begin{document}

\twocolumn[
\icmltitle{AI in Education needs interpretable machine learning: Lessons from Open Learner Modelling}


\icmlsetsymbol{equal}{*}

\begin{icmlauthorlist}
\icmlauthor{Cristina Conati}{UBC}
\icmlauthor{Ka\'{s}ka Porayska-Pomsta}{UCL}
\icmlauthor{Manolis Mavrikis}{UCL}
\end{icmlauthorlist}

\icmlaffiliation{UCL}{University College London, UCL Knowledge Lab, United Kingdom}
\icmlaffiliation{UBC}{University of British Columbia, Department of Computer Science, Canada; }

\icmlcorrespondingauthor{Ka\'{s}ka Porayska-Pomsta}{K.Porayska-Pomsta@ucl.ac.uk}

\icmlkeywords{open learner model, intepretability, machine learning}

\vskip 0.3in
]


\printAffiliationsAndNotice{}

\begin{abstract}
Interpretability of the underlying AI representations is a key raison d'\^{e}tre for Open Learner Modelling (OLM) -- a branch of Intelligent Tutoring Systems (ITS) research. OLMs provide tools for 'opening' up the AI models of learners' cognition and emotions for the purpose of supporting human learning and teaching. Over thirty years of research in ITS (also known as AI in Education) produced important work, which informs about how AI can be used in Education to best effects and, through the OLM research, what are the necessary considerations to make it interpretable and explainable for the benefit of learning. We argue that this work can provide a valuable starting point for a framework of interpretable AI, and as such is of relevance to the application of both knowledge-based and machine learning systems in other high-stakes contexts, beyond education.
\end{abstract}

\section{Introduction of the ITS field}
\label{intro}

Since the early 1970s, the field of Intelligent Tutoring Systems (ITS -- also known as Artificial Intelligence in Education) has been investigating how to leverage AI techniques to create educational technologies that are personalised to the needs of individual learners, with the goal to approximate the well-known benefits of one-to-one instruction ~\citep[for a recent review see][]{Boulay2016}. The idea is essentially to devise intelligent pedagogical agents (IPAs) that can model, predict and monitor relevant learner behaviours, abilities and mental states in a variety of educational activities, and provide personalised help and feedback accordingly \cite{Woolf-2009}. These IPAs need to be able to capture and process information about three main components of the teaching process: (i) the target instructional domain (\emph{domain model}), (ii) the relevant pedagogical knowledge (\emph{pedagogical model}), and (iii) the student themselves (\emph{student model}). These three components define a conceptual architecture of instructional modelling and interaction that emerged from the ITS research over the years ~\citep[e.g.,][]{Boulay2016}. Of these components, the student model constitutes a defining characteristic of an ITS. Non-AI educational technologies, such as test-and-branch systems, generally provide instructional feedback by matching students' responses against some pre-programmed solutions (e.g., \cite{Nesbit2014}). ITS research, on the other hand, strives to provide deeper and more precise pedagogical interventions by modelling in real-time a variety of features that are important for individualised instruction, such as students' domain knowledge and cognitive skills, as well as their meta-cognitive abilities and affective states. Here, it is noteworthy that it is the need for delivering appropriate pedagogical interventions that makes the educational context a high-stakes one for AI: such interventions may have potentially long-lasting impact on peoples learning, development and life-long functioning.

ITS research has successfully delivered techniques and systems that provide personalised support for problem solving activities in a variety of domains (e.g., programming, physics, algebra, geometry, SQL), based on an on-going student modelling assessment of the evolving student domain knowledge during interaction with the system. Formal studies have shown that these systems can foster students' learning better than practicing problem solving skills in class or smaller group contexts, and that the outcomes which are measured in terms of improvements in students' grades are comparable with those achieved through human tutoring ~\cite{Schroeder2013,Nesbit2014,Boulay2016,review_intelligent_2014}.  Some of the ITS are actively used in real-world settings e.g.,\cite{Mitrovic2007,Koedinger2016}, reaching several thousand schools in the US alone and even changing traditional school curricula. The growing shortage of qualified teachers, coupled with growing numbers of students world-wide, represents a substantial societal global challenge and provides a strong motivation to continue investing in ITS research and solutions to enhance and support human learning and development in an accountable way and at scale.

\section{New developments in ITS and the need for machine learning}

For reasons related to the ITS research having started in the tradition of symbolic, rule-based expert systems, as well as owing to the nature of the educational domain where inferential transparency is key to delivering pedagogically effective instruction, a large proportion of ITS is based on knowledge-based techniques. However, there is an emerging appetite and need in the field for machine learning approaches, which is fuelled by a combination of (i) the emergence of big data, e.g. in learning and teaching at scale contexts such as through Massive Open Online Courses (MOOCs), and (ii) a shift within the field towards more complex instructional domains, for which it may be harder to engineer and represent knowledge based on traditional knowledge elicitation from human experts. Specifically, in addition to ITS for problem solving, researchers have been investigating ITSs for a variety of other educational activities in more complex domains that can benefit from individualised support, such as learning from examples ~\cite{Conati2009,LongAleven2017}, learning by exploration or games \cite{ConatiSamad2013,Porayska-Pomsta_SHAREIT_2013,italk2learn2017}, or learning by teaching \cite{Biswas2005}. Providing individualised support for these activities poses unique challenges, because it requires an ITS to model the activities as well as student behaviours, abilities and states that may not be as well-defined, understood or easily captured as those involved in problem solving. For instance, an ITS that provides support for exploration-based learning must be able to ''know'' what it means to explore a given concept or domain effectively (e.g. via an interactive simulation) so that it can monitor the students' exploration process and provide adequate feedback when needed. It might also need to capture and model the domain-independent learner abilities that foster good exploratory behaviour (e.g. self-assessment, planning).

Machine learning (ML) techniques are instrumental in ad- dressing the challenges of these new endeavours, because they can help learn from data the knowledge and models that might be challenging to obtain from human experts, and compute predictions of students' cognitive and mental states in highly dimensional and ill-defined spaces of human behaviours. Examples of ML applications in ITS include modelling student states and abilities such as self-efficacy ~\cite{mavrikis_modelling_2010}, emotional reactions \cite{ConatiMaclaren2009,Bosch2016,Monkaresi2017}, predicting students' ability to successfully conduct scientific inquiries in virtual environments ~\cite{Baker2016}, and automatically generating hints ~\cite{Stamper2011,conati_student_2013,Fratamico2017}. We argue that the interpretability of these techniques, and indeed any other AI techniques employed in an ITS, is critical to enabling an IPA to explain to its users its inferences and actions. The importance of such explanations is two-fold. First, they can improve an IPA's pedagogical effectiveness because they often form an integral part of the skills to be taught (e.g., to help students understand why the system deems their answers to be incorrect or a particular topic to be learned or not).

Second, as in other high-stakes contexts which employ AI for decision-making, an IPA's ability to explain its decisions relates to nurturing users' trust in such decisions \citep[e.g.,][]{MLinHCI2017}. In the ITS context this includes students' trust and their consequent willingness to follow the IPA's suggestions, as well as teachers' trust, which is key for the adoption of these technologies.  ITS researchers have yet to investigate systematically to whom (e.g. student or teacher, or either), why, when and to what extent interpretability and the consequent explainability of an IPA's underlying models can be beneficial. However, in the next section we discuss initial evidence pertaining specifically to the benefits of having interpretable and explainable student models, coming from a branch of ITS research known as Open Learner Modelling (OLM). This research also offers an emergent conceptual framework (outlined in the final section) for understanding the key criteria for interpretable and explainable AI in educational applications. We argue that this conceptual framework along with the examples of different approaches to OLMs may be of relevance to machine learning use in other high-stakes contexts, beyond application in education, in which interpretability, explainability and user control over AI is a requirement.

\section{Open Learner Modelling and Interpretability}

Open Learner Models (OLMs) are student models that allow users to access their content with varying levels of interactivity \cite{Bull1995,Bull2016}. For example, an OLM can be:

\begin{itemize}
\item \emph{scrutable}, i.e. users may view the models current evaluation of relevant student's states and abilities (henceforth -- assessment); 
\item \emph{cooperative} or \emph{negotiable}, where the user and the system work together to arrive at an assessment, 
\item \emph{editable}, namely the user can directly change the model assessment and system's representation of their knowledge at will.
\end{itemize}

Traditionally, OLMs have been designed for students as users of an ITS, with two main purposes: one, pedagogical in nature, is to encourage effective learning skills such as self-assessment and reflection; the second is to improve model accuracy by enabling students to adjust the model's predictions or even its underlying representation when such are deemed inaccurate by the students. Clearly, even OLMs that are merely scrutable require having an underlying representation that is interpretable at some level, so that the model's assessment can be visualised for and understood by its users. However, the more interactive the OLM, the more interpretable and explainable the underlying representations may need to be, because of the increased control that the user has over the access to the different aspects of the model. For example, in a type of negotiation OLM developed by Mabbott and Bull (\citeyear{Mabbott}), the user can register their disagreement with the system's assessment and propose a change. At this point, the system will explain why it believes its current assessment to be correct by providing evidence to support these beliefs, e.g. samples of the learners' previous responses that may indicate a misconception. If the user still disagrees with the system, they have a chance to 'convince' it by answering a series of targeted questions that the system generates. In order to do this, the system keeps a detailed representation of the user's on-task interactions along its assessments of the user's skills.

In the rest of this section we will provide examples of existing OLMs and of their benefits on pedagogical outcomes.

The TARDIS system is an example of ITS that includes a scrutable OLM, namely an OLM that allows students to see the model assessment, but not to interact with the model. TARDIS offers a job interview coaching environment for young people at risk of social exclusion through unemployment. TARDIS includes AI virtual agents which act as recruiters in different job interview scenarios. TARDIS collects evidence from the virtual interviews, based on low-level signals such as the users gaze patterns, gestures, posture, voice activation, etc., and uses machine learning techniques to predict from this evidence the quality of behaviours known to be important for effective interviews (e.g.  appropriate energy in the voice, fluidity of speech, maintenance of gaze with that of the interviewer) \cite{TARDIS2014}.

\begin{figure*}[ht]
\centering
\includegraphics[width=0.80\textwidth]{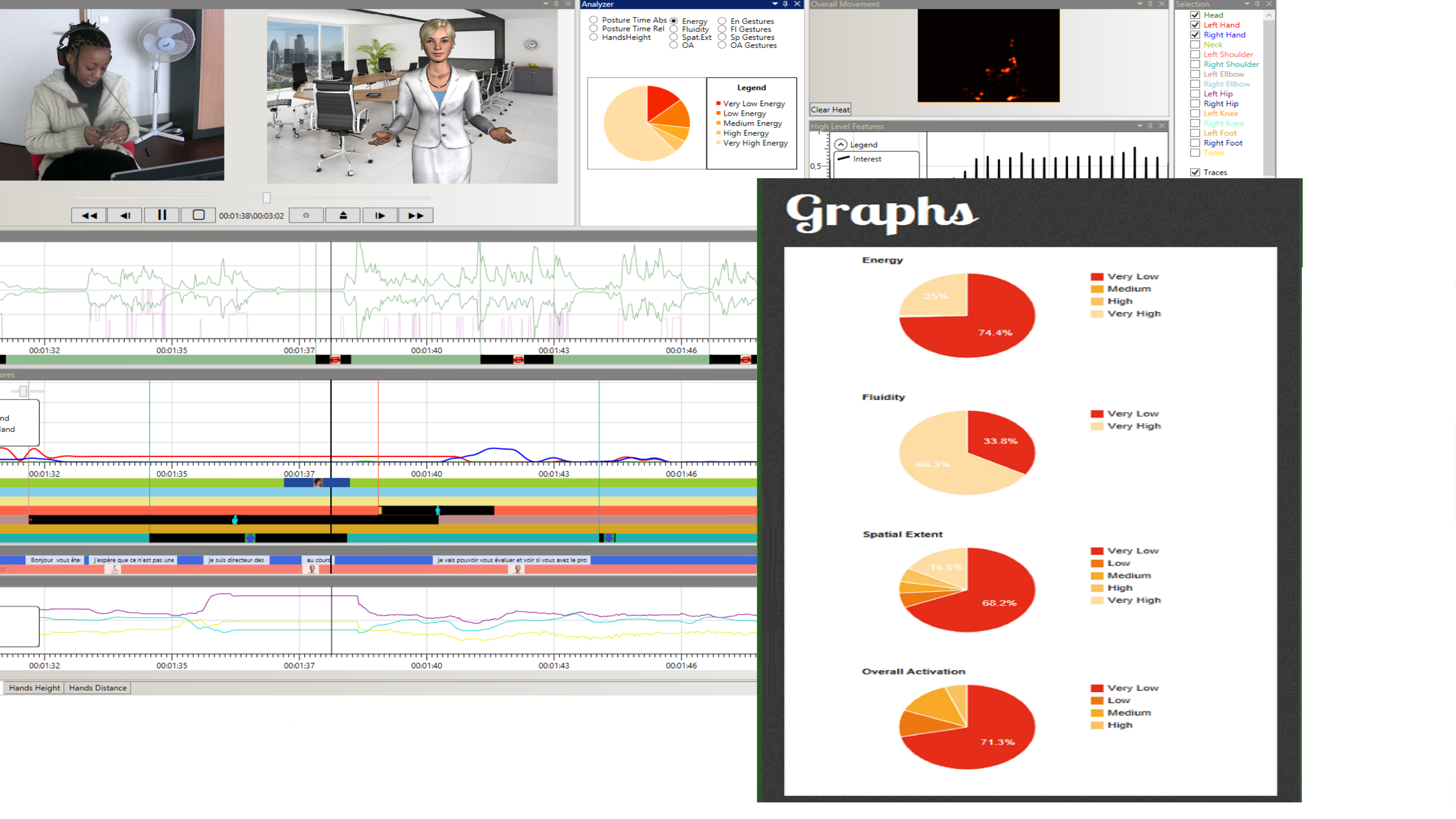}
\caption{TARDIS scrutable OLM showing synchronised recordings of the learners interacting with the AI agents along with the interpretation of the learner's low level social signals such as gaze patterns, gestures, voice activation in terms of higher level judgements about the quality of those behaviours, e.g. energy in voice.}
\label{fig:TARDIS-dashboard}
\end{figure*}  

The model's assessment over these behaviours is then visualised to the learner as shown by the pie charts in Fig. \ref{fig:TARDIS-dashboard}, as a way to provide the users with a concrete and immediate basis for reflecting on how they may improve their verbal and non-verbal behaviours in subsequent interviews with the AI agents. The learner is also shown a time-lined view of the low-level signals and the interpretation thereof.

The information in Fig. \ref{fig:TARDIS-dashboard} is further used by human practitioners to provide nuanced discussion of the learner behaviours in follow-up briefing sessions, showing the importance of interpretable models such as TARDIS's for enhancing human teaching practices. In fact, despite the relatively shallow nature of the information provided by the TARDIS's OLM, a controlled evaluation study with 28 high-risk students, aged 16-18, (14 in OLM and 14 in no OLM condition) showed significant improvements in key behaviours for the OLM condition \cite{Porayska-Pomsta_2018}.

\begin{figure}
\centering
\includegraphics[width=0.50\textwidth]{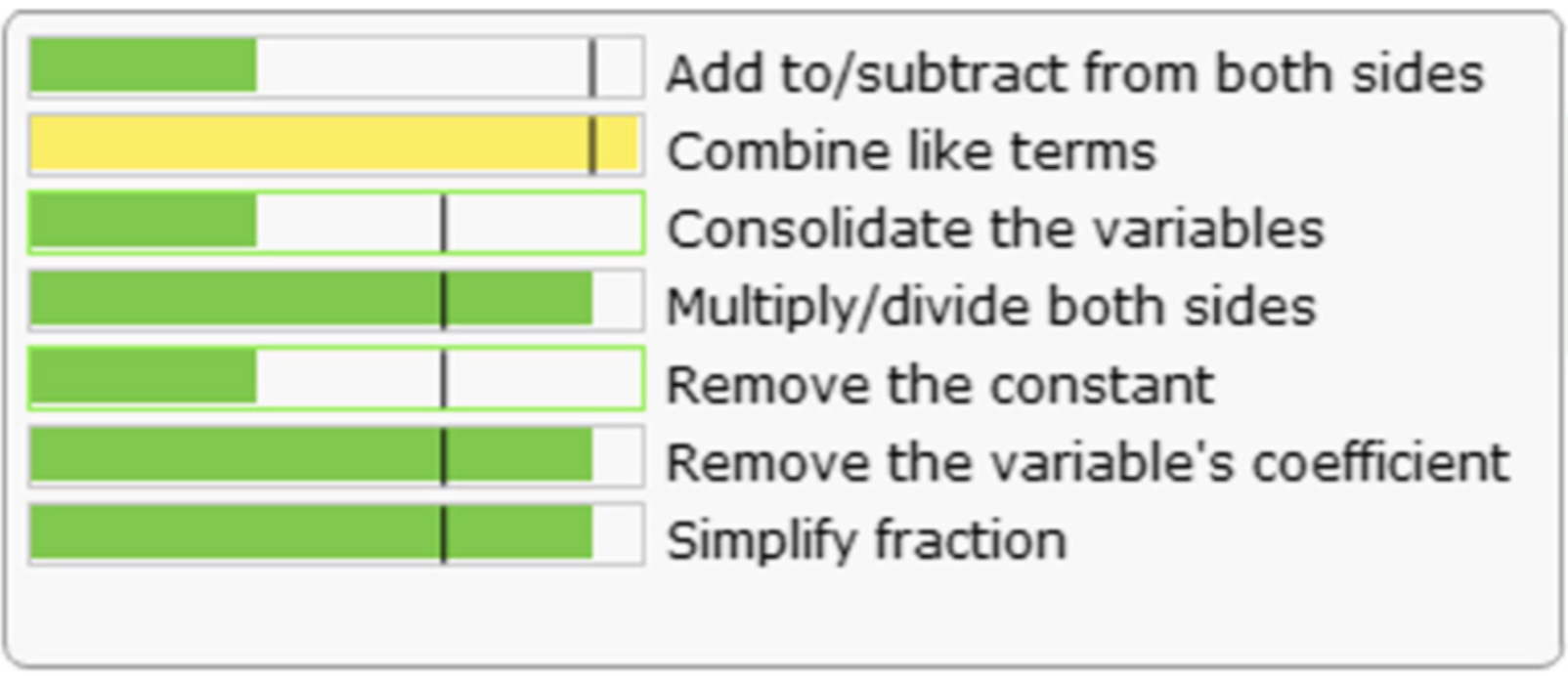}
\caption{Long and Aleven's (2017) skills meter bars indicating the level of student skill mastery}
\label{fig:skillsmeter}
\end{figure}

Our next example is that by \cite{LongAleven2017} where they propose a system that uses a scrutable OLM to help students improve their ability to self-assess their knowledge and share the responsibility for selecting the next problem to work on. The system relies on a student model that is built using the technique known as example-tracing \cite{aleven_example-tracing_2016}: the system evaluates students' problem-solving steps against typical examples of correct problem-solving steps, which are represented in terms of behaviour graphs that encode the different problem-solving strategies applicable for a given problem. Each problem-solving step is related to a piece of domain knowledge (knowledge component, or KC) that needs to be known in order to generate the step. Thus, the evaluation of student's problem-solving steps against the example solutions are used by the system as evidence to generate a probabilistic assessment of student's knowledge (or lack thereof) of the corresponding KC. This process is known as Bayesian Knowledge Tracing (BKT), which has been employed by many ITS to date ~\cite{aleven_knowledge_2013}. The probabilities over KCs generated by BKT are visualised to students in terms of 'skill bars' or skillometer (Fig.1 \ref{fig:skillsmeter}). To support the students' learning and to foster reflection skills, in this OLM the students are asked to self-assess their knowledge of the specific KCs before they can see the system's assessment. The visualisation of such assessment is designed to draw the student's attention to how the assessment changes in response to student's problem-solving actions. That is, once the student asks to see the system's assessment in the skillometer, the relevant bars grow/shrink to new places, based on students' updated skill mastery after finishing the current problem, and as calculated by BKT \cite{Koedinger_Corbett2006}. The dynamic updating of the bars is a form of feedback on students' self-assessment. Here, student self-reflection constitutes an explicit learning goal, which has been shown through a formal user study to significantly improve learning outcomes for the students who used this OLM \cite{LongAleven2017} . 

Our final example is of fully editable OLM, where users have the greatest control over their model. In an early evaluation of user preferences with respect to different forms of OLMs conducted by  \cite{Mabbott}, editable models have been shown to lead to a decreased user trust, especially if the users were novice learners who lacked confidence in their own judgments. More recent examples, however, show that such models can provide effective, engaging and trusted learning tools if they are accompanied by fine-grained support from the system. This in turn necessitates access to detailed model representations and inferences. In a radical approach, ~\citep{Basu2017} implemented an editable OLM which allows students to build models of their knowledge by exploring concepts, properties and relations between them in open ended exploratory learning environments (OELE). To achieve this, they employed a hierarchical representation of tasks and strategies (implemented as a directed acyclic graph) that may need to be undertaken to solve a problem. this representation was that it allowed for the expression of a particular construct or strategy in multiple variations that related to each other, which in turn gave the system access to a set of desired and suboptimal implementations of a task or strategy employed by the user. Based on this, the system can analyse the users behaviour by comparing their action sequences and their associated contexts against desired and suboptimal strategy variants defined in the strategy model and, in turn, to offer targeted support when the users seem to flounder. This representation allows for a conceptual support to be given to the user at a fine-grained level of detail, e.g. low-level objects description in terms of their properties, relations between them and temporal ordering of actions that could be performed on them. This allows the system to guide the user in model building through relatively simple step-by-step interfaces for the different modelling tasks, gradually building users' confidence in their abilities, their buy-in to the system's advice and prompts, ultimately significantly increasing the learning outcomes for the users \cite{Basu2017}.

\section{Discussion and Future Work}
\label{sec:conclusion}

Section 2 established the need for models based on machine learning. As educational technology is deployed at scale, and computational power no longer presents a barrier to adoption, machine learning is used increasingly for cognitive and non-cognitive student modelling. However, to make ML-based models that can be meaningfully employed in the context of supporting human learning and teaching exposes them to demands related to their interpretability, explainability and ultimately trustworthiness ~\cite{WellerWHI2017}. Despite the fact that the vast majority of the OLMs developed over the past thirty years are knowledge-based, the insights offered by the substantial ITS work we introduced here provide an important conceptual and practical framework for developing ML-based models in education, with potential application in other high-stakes contexts concerned with modelling of human behaviour and decision-making.

Initial evaluation studies of the different types of OLMs have started to shed some light on key considerations that need to be taken into account when deciding on both what information to reveal to the user, how and why, as well as to what extent this information needs to approximate the underlying representations of the AI models. The way in which those questions are addressed will have implications for how effective the learning support delivered by an ITS will be.  Comparison studies such as those conducted by ~\cite{Mabbott,KerlyThesis2009} provide initial indications on user preferences for the particular types of OLMs, along with the implications on user trust and improvements in model accuracy that such models engender. For instance, ~\cite{Mabbott} present anecdotal evidence that the maximum level of control facilitated by the editable OLM they tested was the least favoured by the users in their study, compared to non-editable variations of this OLM, because learners did not trust their own judgments and expected targeted support from the system ~\cite{Mabbott}. When such support is not available their trust in the system tends to dwindle along with their motivation to follow the systems instruction. Negotiable or co-operative open learner models that maintain an interaction symmetry where both the system and the learner have to justify and explain their actions, represent the preferred and trusted by users mode for their engagement with such models. Thus, finding the balance between the level of control to be given to the user over the content of their model and the level of system?s support in changing the model that can be offered to them seems critical to deciding the level of algorithmic interpretability needed.

As a summary of the key considerations, we cite four dimensions (expressed as questions) as proposed by \cite{Bull2016}:
\begin{enumerate}
\item \emph{Why} the OLM is being built, e.g. to improve model accuracy, to engender user right of access and trust, to nurture self-regulation and reflection, etc.;
\item \emph{Which} aspects of the model are made available to the user. Examples include the extent of the learner model that is open; closeness of the externalisation of the learner model to the underlying model representations; extent of access to (un)certainty in the model's assessment; access to different temporal views, e.g. current, past, anticipated future models; access to sources of input to the model; access to explanation of the relationship between the learner model and personalisation of the interventions based on such a model;
\item \emph{How} is the model accessed and the degree to which it can be manipulated by the user, such as visualisation used in the OLM; type of interactivity with the model; flexibility of access to the model;
\item \emph{Who} has access to the model, e.g. intended users such as students, teachers, parents.  
\end{enumerate}

The four dimensions allow OLM architects to calibrate, at least in principle, pedagogically optimal designs of those tools. Much research is still needed to deliver a universal framework for interpretable AI and to understand better how to make ML-based OLMs viable in supporting human learning and development at scale. Nevertheless, we propose that the examples and the preliminary empirical findings gleaned from the work on OLMs in the context of Artificial Intelligence in Education have implications for how we may address the need for interpretability of AI models, be it knowledge- or ML-based, and offer additional, and thus far, largely ignored conceptual starting point from ITS research for consideration by a wider AI and machine learning community.

\bibliography{WHI-OLM2}
\bibliographystyle{icml2018}

\end{document}